\documentclass{article}

\usepackage[preprint]{neurips_2019}

\usepackage[utf8]{inputenc} 
\usepackage[T1]{fontenc}    
\usepackage{hyperref}       
\usepackage{url}            
\usepackage{booktabs}       
\usepackage{amsfonts}       
\usepackage{nicefrac}       
\usepackage{microtype}      

\usepackage{amsmath}
\usepackage{subfig}
\usepackage{enumitem}
\usepackage{float}
\usepackage{algorithm}
\usepackage{algorithmic}
\usepackage{graphicx}
\usepackage{bbm}
\usepackage{multirow}

\newcommand{\e}{\epsilon}
\newcommand{\y}{\gamma}
\newcommand{\al}{\alpha}

\newcommand{\ta}{\theta}
\newcommand{\w}{\omega}
\newcommand{\E}{\mathbb{E}}

\newcommand{\lp}{\left (}
\newcommand{\rp}{\right )}
\newcommand{\B}{\mathcal{B}}
\newcommand{\Loss}{\mathcal{L}}

\DeclareMathOperator*{\argmin}{argmin}
\DeclareMathOperator*{\argmax}{argmax}

\title{Benchmarking Batch Deep Reinforcement Learning Algorithms}

\author{
Scott Fujimoto$^{1,2}$, Edoardo Conti$^2$, Mohammad Ghavamzadeh$^3$, Joelle Pineau$^{1,3}$\\
$^1$Mila, McGill University\\
$^2$Facebook\\
$^3$Facebook AI Research\\
\texttt{scott.fujimoto@mail.mcgill.ca}
}

\begin{document}

\maketitle

\begin{abstract}
Widely-used deep reinforcement learning algorithms have been shown to fail in the batch setting--learning from a fixed data set without interaction with the environment. Following this result, there have been several papers showing reasonable performances under a variety of environments and batch settings. In this paper, we benchmark the performance of recent off-policy and batch reinforcement learning algorithms under unified settings on the Atari domain, with data generated by a single partially-trained behavioral policy. We find that under these conditions, many of these algorithms underperform DQN trained online with the same amount of data, as well as the partially-trained behavioral policy. To introduce a strong baseline, we adapt the Batch-Constrained Q-learning algorithm to a discrete-action setting, and show it outperforms all existing algorithms at this task.
\end{abstract}

\section{Introduction}

Batch reinforcement learning is the study of algorithms that can learn from a single batch of data, without directly interacting with the environment \citep{lange2012batch}. Learning with finite data sets is invaluable for a variety of real-world applications, where data collection may be difficult, time-consuming or costly \citep{guez2008adaptive, pietquin2011sample, gauci2018horizon}. In principle, standard off-policy deep reinforcement learning algorithms such as DQN and DDPG \citep{DQN, DDPG} are applicable in the batch reinforcement learning setting, due to basis on more fundamental batch reinforcement learning algorithms such as Fitted Q-iteration \citep{ernst2005tree, riedmiller2005neural}. However, these traditional algorithms only come with convergence guarantees for non-parametric function approximation \citep{gordon1995stable, ormoneit2002kernel}, have no guarantees on the quality of the learned policy, and scale poorly to high dimensional tasks.

Recent results demonstrated widely-used off-policy deep reinforcement learning algorithms fail in the batch setting due to a phenomenon known as extrapolation error, which is induced from evaluating state-action pairs which are not contained in the provided batch of data \citep{fujimoto2019off}. This erroneous extrapolation is propagated through temporal difference update of most off-policy algorithms \citep{sutton1988tdlearning}, causing extreme overestimation and poor performance \citep{thrun1993bias}. \cite{fujimoto2019off} proposed the batch-constrained reinforcement learning framework, where the agent should favor a state-action visitation similar to some subset of the provided batch, and provided a practical continuous control deep reinforcement learning algorithm, BCQ, which eliminates unseen actions through a sampling procedure over a generative model of the data set. Contrary to these results, \cite{agarwal2019striving} showed that using the entire history of a deep reinforcement learning agent as a batch (50 million time steps), standard deep reinforcement learning algorithms could reach a comparable performance to an online algorithm. In particular, they highlighted that distributional reinforcement learning algorithms \citep{bellemare2017distributional, dabney2017distributional} performed particularly well with this large and diverse data set. 

Given batch reinforcement learning encompasses a large number of settings, existing algorithms have been tested over a wide range of environments, and under a variety of data distributions, making comparisons difficult and contradictory results possible. In this paper, we benchmark the performance of several algorithms on the Arcade Learning Environment \citep{bellemare2013arcade}, with a large data set of 10 million data points generated by a partially-trained policy. 
We find that with a single behavioral policy, 
not only do widely-used off-policy deep reinforcement learning algorithms perform poorly, even existing batch algorithms are inadequate solutions, failing to outperform the behavioral policy.

We introduce a variant of the BCQ algorithm \citep{fujimoto2019off} which operates on a discrete action space. Our version of BCQ is simple to implement, while maintaining the core ideas of the original continuous control algorithm. Furthermore, our results demonstrate BCQ greatly outperforms all prior deep batch reinforcement learning algorithms, including KL-Control \citep{jaques2019way}, which was shown to outperform another na\"ive variant of BCQ for discrete actions. BCQ demonstrates learning akin to a strong robust imitation learning algorithm, matching, or exceeding, the performance of the noiseless behavioral policy, a DQN agent trained online with the same amount of data. While simply matching the noiseless behavioral policy is often unsatisfactory, we hope that BCQ will serve as a strong baseline in this setting.

Our contributions are as follows: 
\begin{itemize}
	\item We benchmark the performance of several batch deep reinforcement learning algorithms under a single unified setting. This continues the line of work from \cite{agarwal2019striving} by examining the performance of widely-used off-policy algorithms in the Atari domain. However under ordinary data conditions, we find that standard off-policy reinforcement learning algorithms perform poorly.
    \item We validate the batch reinforcement learning experiments from \cite{fujimoto2019off} on the more challenging, discrete-action Atari environments, and demonstrate the phenomenon of extrapolation error still occurs in this domain. 
    \item We introduce a discrete-action version of BCQ which achieves a state of the art performance in our batch reinforcement learning setting, and will serve as a strong baseline for future methods.
\end{itemize}

\section{Preliminaries}

\textbf{Reinforcement Learning.} Reinforcement learning studies sequential decision making processes, generally formulated by a Markov decision process (MDP) $(\mathcal{S}, \mathcal{A}, p, r, \y)$, where $\mathcal{S}$ and $\mathcal{A}$ denote the corresponding state and action spaces respectively. 
At a given discrete time step, a reinforcement learning agent takes action $a \in \mathcal{A}$ in state $s \in \mathcal{S}$, and receives a new state $s'  \in \mathcal{S}$ and reward $r(s,a,s')$, in accordance to the transition dynamics $p(s',r|s,a)$. The aim of the agent is to maximize the sum of discounted rewards, also known as the return $R_t = \sum_{i={t+1}}^\infty \y^i r(s_i, a_i, s_{i+1})$, where the discount factor $\y \in [0,1)$, determines the effective horizon by weighting future rewards. The decisions of an agent are made by its policy $\pi: \mathcal{S} \rightarrow \mathcal{A}$, which maps a given state $s$ to a distribution over actions.

For a given policy $\pi$, we define the value function as the expected return of an agent following the policy $Q^\pi(s,a)=\E_\pi[R_t|s,a]$. Given $Q^\pi$, a new policy $\pi'$ of equal or better performance can be derived by greedy maximization $\pi' = \argmax_a Q^\pi(s,a)$~\citep{sutton1998reinforcement}. The optimal policy $\pi^* = \argmax Q^*(s,a)$, can be obtained by greedy selection over the optimal value function $Q^*(s,a) = \max_\pi Q^\pi(s,a)$. For $\y \in [0,1)$ the value function and optimal value function are the unique fixed points of the Bellman operator $\mathcal{T}^\pi$ and optimality operator $\mathcal{T}^*$, respectively \citep{bellman, bertsekas1995dynamic}:
\begin{align}
\mathcal{T}^\pi Q(s,a) &= \E_{s', r, a'\sim \pi} [r + \y Q(s',a')] \\
\mathcal{T}^* Q(s,a) &= \E_{s', r} [r + \y \max_{a'} Q(s',a')].
\end{align}

\textbf{Deep Reinforcement Learning.} In deep reinforcement learning, the value function is approximated by a neural network $Q_\ta$. In the Deep Q-Network algorithm (DQN) \citep{DQN}, this value function $Q_\ta$ is updated in a manner that approximates the optimality operator, through Q-learning~\citep{watkins1989qlearning}:
\begin{equation} \label{eqn:DQN}
    \Loss(\ta) = l_\kappa \lp r + \gamma \max_{a'} Q_{\ta'}(s',a') - Q_\ta(s,a) \rp,
\end{equation}
where $l_\kappa$ defines the Huber loss \citep{huber1964robust}:     
\begin{equation}
    l_\kappa(\delta) = 
    \begin{cases}
    0.5 \delta^2 &\text{if } \delta \leq \kappa\\
    \kappa(|\delta| - 0.5\kappa) &\text{otherwise.}
    \end{cases},
\end{equation}
but is generally interchangeable with other losses such as mean-squared error. A target network $Q_{\ta'}$ with frozen parameters is used to maintain a fixed target over multiple updates, where $\ta'$ is updated to $\ta$ after a set number of learning steps. The loss (Eqn. \ref{eqn:DQN}) is minimized over mini-batches of transitions $(s,a,r,s')$ sampled from some data set, or replay buffer $\B$ \citep{expreplay1992}. For an \textit{on-policy} algorithm, $\B$ is generated by the current policy, however, for an \textit{off-policy} algorithm $\B$ may be generated by any collection of policies.

\textbf{Batch Deep Reinforcement Learning.} In \textit{batch reinforcement learning}, we additionally assume the data set is fixed, and no further interactions with the environment will occur. This is in contrast to many off-policy deep reinforcement learning algorithms which assume further interactions with the current policy, but train with a history of experiences generated by previous iterations of the policy. In some instances, access to the behavioral policy is assumed \citep{precup2001off, thomas2016data, petrik2016safe, laroche2019safe}, but in our experiments, the behavioral policy is treated as unknown. For notational simplicity, we sometimes refer to a collection of behavioral policies as a single behavioral policy $\pi_b$.

Batch deep reinforcement learning algorithms have been shown to be susceptible to \textit{extrapolation error} \citep{fujimoto2019off}, induced by generalization from the neural network function approximator. When selecting actions $a'$ (Eqn. \ref{eqn:DQN}), such that $(s',a')$ is distant from data contained in the batch, the estimate $Q_{\ta'}(s',a')$ may be arbitrarily poor, introducing extrapolation error. In systems where further environment interactions are possible this error can be mitigated by simply attempting the action $a'$, which occurs naturally as long as the behavioral policy is similar to the target policy. 

\section{Batch Deep Reinforcement Learning Algorithms} \label{section:brl}

In this section, we survey recent batch deep reinforcement learning algorithms, including off-policy algorithms which have been to shown to work in a batch setting \citep{agarwal2019striving}. 

\textbf{QR-DQN.} Quantile Regression DQN (QR-DQN) \citep{dabney2017distributional} is a distributional reinforcement learning method \citep{morimura2010nonparametric, bellemare2017distributional} which aims to estimate the set of $K$ $\tau$-quantiles of the return distribution, $\{\tau\}^K = \{\frac{i + 0.5}{K}\}^{K-1}_{i=0}$. Instead of outputting a single value for each action, QR-DQN outputs a $K$-dimensional vector representing these quantiles. A pairwise loss between each quantile is computed, similarly to DQN: 
\begin{equation}
    \Loss(\ta) = \frac{1}{K^2} \sum_\tau \sum_{\tau'} l_\tau \lp r + \gamma \max_{a'} Q^{\tau'}_{\ta'}(s',a') - Q^\tau_\ta(s,a) \rp,    
\end{equation}
where $l_\tau$ is a weighted variant of the Huber loss, denoted the quantile Huber loss:
\begin{equation}
    l_\tau(\delta) = |\tau - \mathbbm{1}_{\delta < 0}|l_\kappa(\delta).
\end{equation}
An estimate of the value can be recovered through the mean over the quantiles, and the policy $\pi$ is defined by greedy selection over this value:
\begin{equation}
\pi(s) = \argmax_a \frac{1}{K} \sum_\tau Q^\tau_\ta(s,a).
\end{equation}

\textbf{REM.} Random Ensemble Mixture (REM) \citep{agarwal2019striving} is an off-policy Q-learning method which aims to capture the success of distributional reinforcement learning algorithms with a simpler algorithm. Similar to QR-DQN, the output of the Q-network is a $K$-dimensional vector. During each update, this vector is combined with a convex combination of $K$ weights $\al_k$ sampled from a $(K-1)$-simplex: 
\begin{equation}
    \Loss(\ta) = l_\kappa \lp r + \gamma \max_{a'} \sum_k \al_k Q^k_{\ta'}(s',a') - \sum_k \al_k Q^k_\ta(s,a) \rp.
\end{equation}
The policy is defined by the argmax over the mean of the output vector $\pi = \argmax_a \frac{1}{K} \sum Q^k_\ta(s,a)$.

\textbf{BCQ.} Batch-Constrained deep Q-learning (BCQ) \citep{fujimoto2019off} is a batch reinforcement learning method for continuous control. 
BCQ aims to perform Q-learning while constraining the action space to eliminate actions which are unlikely to be selected by the behavioral policy $\pi_b$, and are therefore unlikely to be contained in the batch.
At its core, BCQ uses a state-conditioned generative model $G_\w: \mathcal{S} \rightarrow \mathcal{A}$ to model the distribution of data in the batch, $G_\w \approx \pi_b$ akin to a behavioral cloning model. As it is easier to sample from $\pi_b(a|s)$ than model $\pi_b(a|s)$ exactly in a continuous action space, the policy is defined by sampling $N$ actions $a_i$ from $G_\w(s)$ and selecting the highest valued action according to a Q-network. 
Since BCQ was designed for continuous actions, the method also includes a perturbation model $\xi_\phi(s,a)$, which is a residual added to the sampled actions in the range $[-\Phi, \Phi]$, and trained with the deterministic policy gradient \citep{DPG}. Finally the authors include a weighted version of Clipped Double Q-learning \citep{fujimoto2018addressing} to penalize high variance estimates and reduce overestimation bias, using $Q^k_\ta$ with $k=\{1,2\}$:
\begin{align}
\Loss(\ta) &= \sum_{k} \lp r + \y \max_{\hat a} \lp \lambda \min_{k'} Q^{k'}_{\ta'}(s',\hat a) + (1 - \lambda) \max_{k'} Q^{k'}_{\ta'}(s',\hat a) \rp - Q^k_\ta(s,a) \rp ^2, \\
&\text{where } \hat a = a_i + \xi_\phi(s',a_i), \qquad a_i \sim G_\w(s'). \nonumber 
\end{align}
During evaluation, the policy is defined similarly, by sampling $N$ actions from the generative model, perturbing them and selecting the argmax:
\begin{equation}
\pi(s) = \argmax_{\hat a = a_i + \xi_\phi(s',a_i)} Q^0_\ta(s,\hat a), \qquad a_i \sim G_\w(s).
\end{equation}

\textbf{BEAR-QL.} Bootstrapping Error Accumulation Reduction Q-Learning (BEAR-QL) \citep{kumar2019stabilizing} is an actor-critic algorithm which builds on the core idea of BCQ, but instead of using a perturbation model, samples actions from a learned actor. 
As in BCQ, BEAR-QL trains a generative model of the data distribution in the batch. Using the generative model $G_\w$, the actor $\pi_\phi$ is trained using the deterministic policy gradient \citep{DPG}, while minimizing the variance over an ensemble of $K$ Q-networks, and constraining the maximum mean discrepancy (MMD) \citep{gretton2012kernel} between $G_\w$ and $\pi_\phi$ through dual gradient descent: 
\begin{equation}
\Loss(\phi) = -\lp \frac{1}{K} \sum_k Q^k_\ta(s,\hat a) - \tau \text{var}_k Q^k_\ta(s,\hat a) \rp \text{ s.t. MMD}\lp G_\w(s), \pi_\phi(s)\rp \leq \e,
\end{equation}
where $\hat a \sim \pi_\phi(s)$ and the MMD is computed over some choice of kernel. The update rule for the ensemble of Q-networks matches BCQ, except the actions $\hat a$ can be sampled from the single actor network $\pi_\phi$ rather than sampling from a generative model and perturbing: 
\begin{equation}
\Loss(\ta) = \sum_k \lp r + \y \max_{\hat a \sim \pi_{\phi}(s')} \lp \lambda \min_{k'} Q^{k'}_{\ta'}(s',\hat a) + (1 - \lambda) \max_{k'} Q^{k'}_{\ta'}(s',\hat a)\rp - Q^k_\ta(s,a) \rp^2.
\end{equation}
The policy used during evaluation is defined similarly to BCQ, but again samples actions directly from the actor: 
\begin{equation}
\pi(s) = \argmax_{\hat a} \frac{1}{K} \sum_k Q^k_\ta(s,\hat a), \qquad \hat a \sim \pi_\phi(s).
\end{equation}

\textbf{KL-Control.} KL-Control with $\Psi$-learning and Monte-Carlo target value estimation is a combination of methods introduced by \cite{jaques2019way} for batch reinforcement learning of a dialog task with discrete actions. KL-control uses a KL-regularized objective to incorporate a prior $p$ into learning, which is weighted by the hyper-parameter $c$. In this method, the prior is set to a learned estimate of the behavioral policy $p=\pi_b$, again via a generative model $G_\w$, in similar fashion to BCQ, noting that in a discrete-action setting the probabilities $G_\w(a|s) \approx \pi_b(a|s)$ can be computed exactly through behavioral cloning. Rather than a hard maximum in the target, $\Psi$-learning uses the log over the sum of the exponential of the action-values \citep{jaques2017sequence}. Finally, \cite{jaques2019way} estimate a lower-bound over the target value by Monte-Carlo estimation, from sampling $K$ dropout masks \citep{srivastava2014dropout, gal2016dropout} and taking the minimum: 
\begin{equation}
    \Loss(\ta) = l_\kappa \lp \log G_\w(a|s) + \frac{r}{c} + \gamma \min_k \lp \log \sum_{a'} \exp{Q^k_{\ta'}(s',a')} \rp - Q_\ta(s,a) \rp.
\end{equation}
In $\Psi$-learning, the policy $\pi(a|s) = \frac{\exp{Q_\ta(s,a)}}{\sum_{\hat a \in \mathcal{A}} \exp Q_\ta(s,\hat a)}$, as a form of Boltzmann exploration, however in a setting where diversity is not beneficial, we found $\pi = \argmax_a Q_\ta(s,a)$ to achieve higher performance. 

\textbf{SPIBB-DQN.} Safe Policy Improvement with Baseline Bootstrapping DQN (SPIBB-DQN) \citep{laroche2019safe} is a safe batch reinforcement learning algorithm for discrete actions which resembles Q-learning, but modifies the policy to match the behavioral policy, or a known baseline, $\pi_b$ when there is insufficient access to data. Additionally, the authors assume access to some estimate of the state-action distribution of the batch $D(s,a)$. The authors define state-action pairs $(s,a) \in \mathfrak{B}$ as the data points which are unlikely under the data distribution $D(s,a) \leq \e$. For a given state $s$, the actions $a$, such that $(s,a) \in \mathfrak{B}$, the policy is set to match the baseline policy $\pi(a|s) = \pi_b(a|s)$. Otherwise, the highest valued action $a \notin \mathfrak{B}$ is set to the remaining probability, defining the policy as follows:
\begin{equation}
\pi(a|s) = 
\begin{cases}
\pi_b(a|s) &\text{if } (s,a) \in \mathfrak{B} \\
\sum_{a \notin \mathfrak{B}} \pi_b(a|s) &\text{if } (s,a) \notin \mathfrak{B} \text{ and } a = \argmax_{a | (s,a) \notin \mathfrak{B}} Q_\ta(s,a) \\
0 &\text{otherwise.}
\end{cases}.
\end{equation}
The Q-network is updated following the standard DQN update, swapping the max operator with $\pi$:
\begin{equation}
\Loss(\ta) = l_\kappa \lp r + \gamma Q_{\ta'}(s',a') - Q_\ta(s,a) \rp, \qquad a' \sim \pi(s').
\end{equation}
Although appealing due to its theoretical guarantees in the tabular setting, the authors unfortunately do not include a complete implementation of their deep algorithm in their paper, instead analyzing the algorithm under settings where $D(s,a)$ can be computed almost exactly. However, in principle $D(s,a)$ can be computed with a number of pseudo-count methods~\citep{bellemare2016unifying, tang2017exploration, burda2018exploration}.

\section{Discrete Batch-Constrained Deep Q-learning}

In this section, we introduce a discrete variant of the Batch Constrained deep Q-Learning (BCQ) algorithm \citep{fujimoto2019off}. Much of the complexity of the original algorithm is introduced to deal with the continuous action space, and the core principles of the algorithm can be maintained in a much simpler manner in the discrete setting. 

In the original BCQ, a state-conditioned model of the data set $G_\w$ is learned and the highest valued action is selected after sampling actions from $G_\w$ and perturbing the sampled actions within a set range. However, in a discrete-action setting, we can compute the probabilities of every action $G_\w(a|s) \approx \pi_b(a|s)$, and instead utilize some threshold to eliminate actions:
\begin{equation}
    \pi(s) = \argmax_{a | G_\w(a|s) / \max \hat a~G_\w(\hat a|s) > \tau } Q_\ta(s,a).
\end{equation}
To adaptively adjust this threshold, we scale it by the maximum probability from the generative model over all actions, to only allow actions whose relative probability is above some threshold. This results in an algorithm comparable to DQN \citep{DQN} where the policy is defined by a constrained argmax. The Q-network is trained by swapping the max operation with actions selected by the policy:
\begin{equation}
    \Loss(\ta) = l_\kappa \lp r + \y \max_{a'| G_\w(a'|s') / \max \hat a~G_\w(\hat a|s') > \tau} Q_{\ta'}(s',a') - Q_\ta(s,a) \rp.
\end{equation}
With this threshold $\tau$, we maintain the original property of BCQ where setting $\tau=0$ returns Q-learning and $\tau=1$ returns an imitator of the actions contained in the batch.

Given the original BCQ used Clipped Double Q-learning \citep{fujimoto2018addressing} to reduce overestimation bias in the continuous-action setting, we instead apply Double DQN \citep{DoubleDQN}, selecting the max valued action with the current Q-network $Q_\ta$, and evaluating with the target Q-network $Q_{\ta'}$:
\begin{equation}
    \Loss(\ta) = l_\kappa \lp r + \y Q_{\ta'}(s',a') - Q_\ta(s,a) \rp, \qquad a' =\argmax_{a'| G_\w(a'|s') / \max \hat a~G_\w(\hat a|s') > \tau} Q_{\ta}(s',a').
\end{equation}

The generative model $G_\w$, effectively a behavioral cloning network, is trained in a standard supervised learning fashion, with a cross-entropy loss. We summarize our discrete BCQ in algorithm \ref{algorithm:BCQ}.

\begin{algorithm}[t]
  \caption{BCQ}
  \label{algorithm:BCQ}
\begin{algorithmic}[1]
	\STATE \textbf{Input:} Batch $\B$, number of iterations $T$, target\_update\_rate, mini-batch size $N$, threshold $\tau$. 
	\STATE Initialize Q-network $Q_\ta$, generative model $G_\w$ and target network $Q_{\ta'}$ with $\ta' \leftarrow \ta$.
	\FOR{$t=1$ {\bfseries to} $T$}
	\STATE Sample mini-batch $M$ of $N$ transitions $(s, a, r, s')$ from $\B$.
	\STATE $a' =\argmax_{a'| G_\w(a'|s') / \max \hat a~G_\w(\hat a|s') > \tau} Q_{\ta}(s',a')$
	\STATE $\ta \leftarrow \argmin_\ta \sum_{(s,a,r,s') \in M} l_\kappa \lp r + \y Q_{\ta'}(s',a') - Q_\ta(s,a) \rp$
	\STATE $\w \leftarrow \argmin_\w -\sum_{(s,a) \in M} \log G_\w(a|s)$ 
	\STATE If $t \text{ mod target\_update\_rate} = 0$: $\ta' \leftarrow \ta$
	\ENDFOR
\end{algorithmic}
\end{algorithm}

\section{Experiments}

\begin{figure}
\centering
\includegraphics[width=\linewidth]{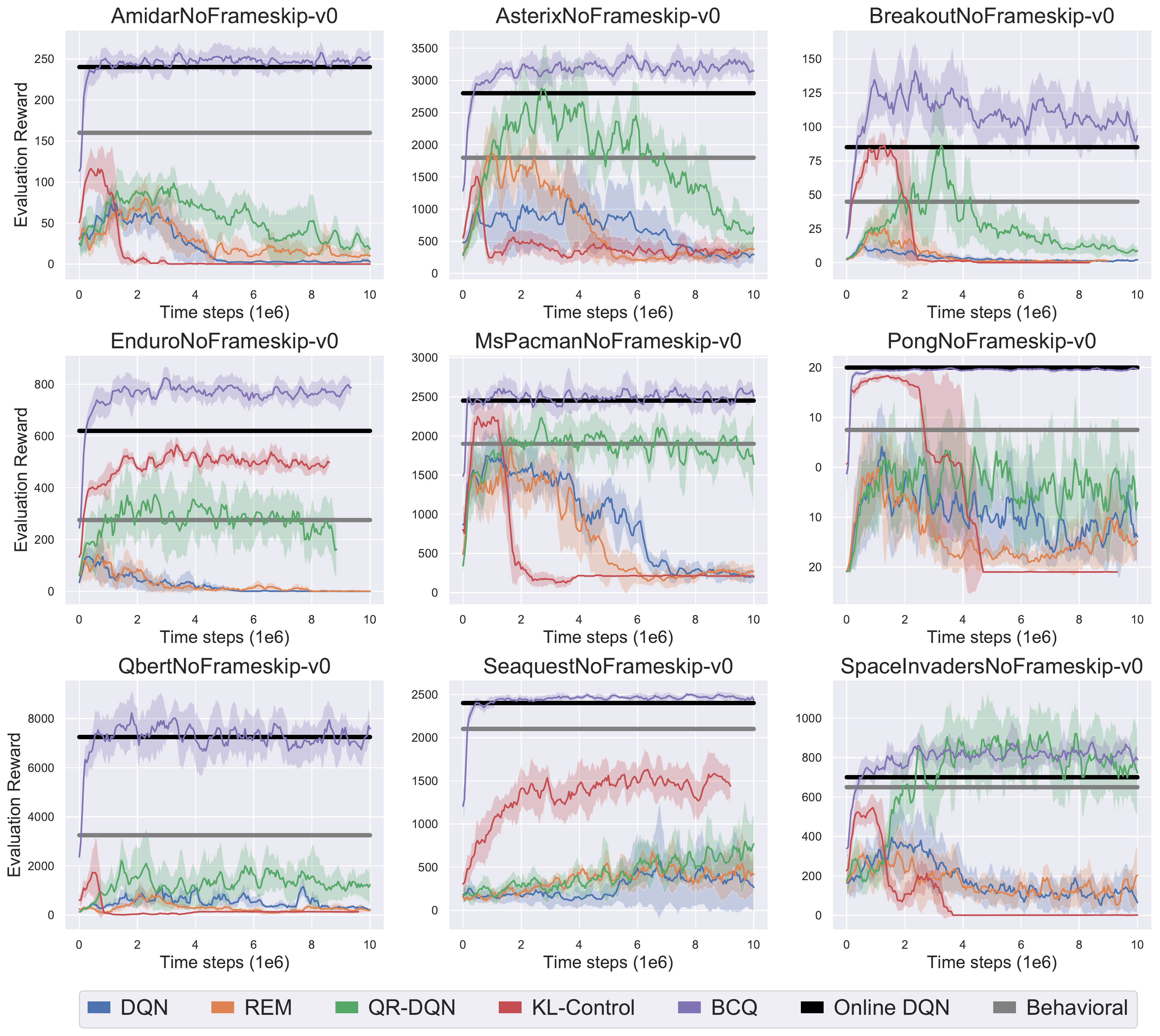}
\caption{Results on 9 Atari 2600 games. Agents are trained from a buffer of 10 million transitions collected by a single partially-trained DQN. Performance of online DQN, and the behavorial policy (online DQN with added noise) is included. 
Agents are evaluated every 50k time steps over 10 episodes, and averaged over 3 seeds and a sliding window of 5. The shaded area measures a single standard deviation across trials.}
\label{fig:results}
\end{figure}

\begin{figure}
\centering
\includegraphics[width=\linewidth]{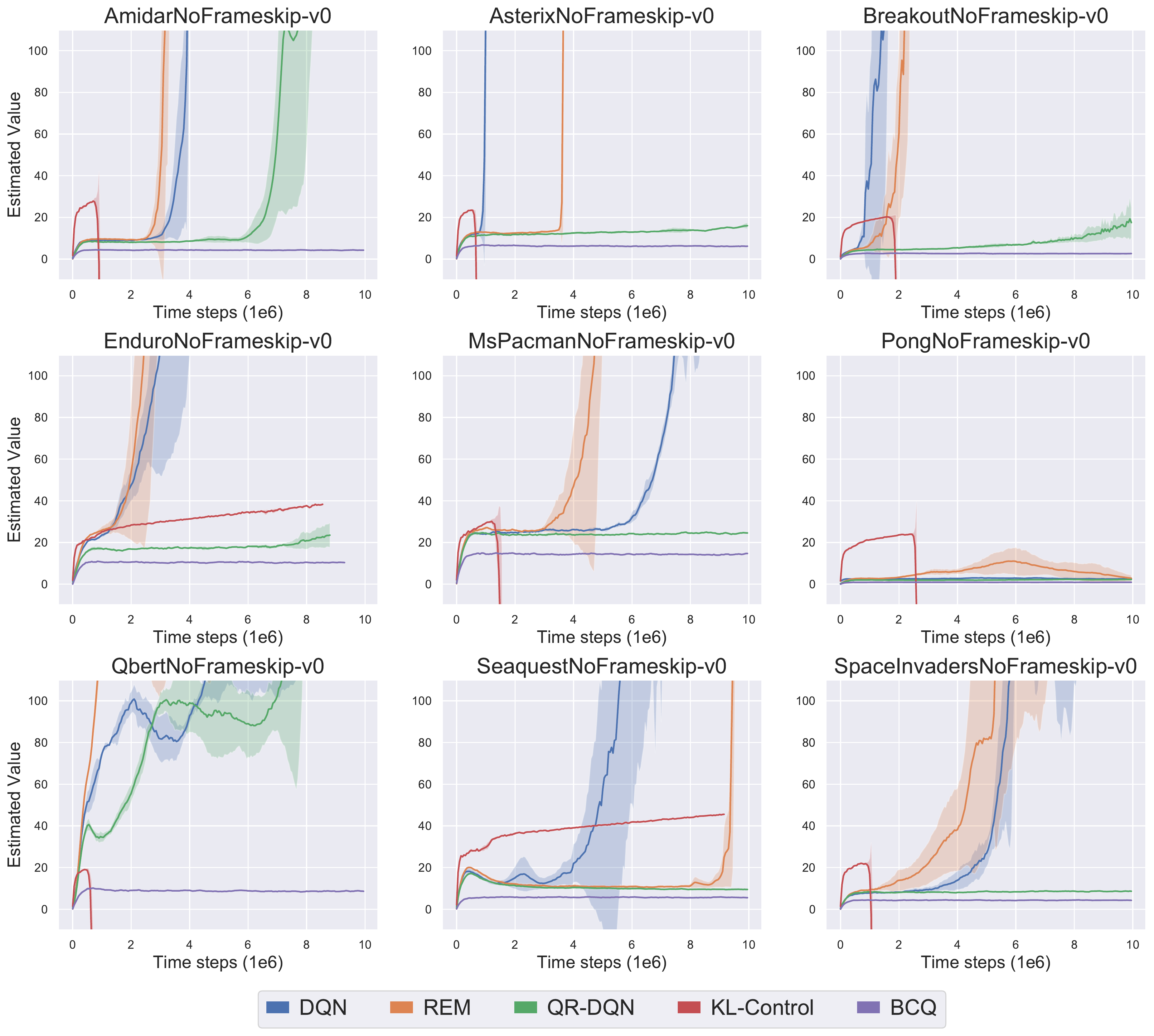}
\caption{Value estimates from the Q-networks of each agent on 9 Atari 2600 games. This figure shows which agents have stable outputs, demonstrating a resistance to extrapolation error. Additionally, drops in performance (Figure \ref{fig:results}) can be seen to correspond to divergence in the value estimates. Value estimates are averaged over 5000 mini-batches of 32 and 3 seeds. The shaded area measured a single standard deviation across trials, clipped to 100 for visual clarity.}
\label{fig:values}
\end{figure}

We validate our methods on the Arcade Learning Environment platform \citep{bellemare2013arcade} of Atari 2600 games through OpenAI gym \citep{OpenAIGym}. We use standard preprocessing to image frames and environment rewards \citep{DQN, castro2018dopamine}, and match the recommendations of \cite{machado2018revisiting} for fair and reproducible results. Exact experimental and algorithmic details are explained in the supplementary. In general, consistent hyper-parameters are kept across all algorithms, and minimal hyper-parameter optimization was performed. 

\textbf{Algorithms.} We use each of the algorithms listed in Section \ref{section:brl}, other than BEAR-QL and SPIBB-DQN, and use our discrete version of BCQ rather than the original, which was defined for continuous control. We omit BEAR-QL due to the reliance on a continuous action space, similarity to BCQ and number of hyper-parameter choices. Although SPIBB-DQN can be extended to settings where the behavioral policy is estimated \citep{simao2019safe}, we omit it due to the lack of implementation for a deep setting without access to pseudo-counts of the dataset, which again would require a large number of design and hyper-parameter choices. 


\textbf{Experimental Setting.} We use a partially-trained DQN agent \citep{DQN} as our behavioral policy. This DQN agent was trained online for 10 million time steps (40 million frames), following a standard training procedure. The behavioral policy is used to gather a new set of 10 million transitions which is used to train each off-policy agent. To ensure exploration, the behavioral policy uses $\e = 0.2$ for the whole episode with $p=0.8$ and $\e=0.001$ with $p=0.2$. This mix of $\e$ is used to ensure the data set includes data reaching the max performance of the agent as well as exploratory behavior. Unlike the experiments by \cite{agarwal2019striving}, the batch data is generated by this single behavioral policy, rather than a series of changing policies. This setup is used to closely match batch settings used by real-world systems, which generally rely on a single behavioral for a fixed period of time \citep{gauci2018horizon}. For each environment, the agents are trained on this data set for 10 million time steps, and evaluated on 10 episodes every 50k time steps. We graph both the final performance of the online DQN, as well as the performance of the behavioral policy, the online DQN with exploration noise. Results are displayed in Figure \ref{fig:results}. Additionally, in Figure \ref{fig:values} we graph the value estimates of each algorithm to examine if the divergence from extrapolation error \citep{fujimoto2019off} is present in the Atari domain.

\textbf{Discussion.} Under our experimental conditions, both the online DQN and offline agents have been trained with $10$ million data points, where the offline agents are trained with $4 \times$ more iterations. Regardless, it is clear from Figure \ref{fig:results} that standard off-policy algorithms (DQN, QR-DQN, REM) perform poorly in this single behavioral policy setting. Out of the three QR-DQN is a clear winner, but generally underperforms the noisy behavioral policy. While \cite{agarwal2019striving} showed these algorithms performed well with large replay buffers and high diversity, by training agents using the entire replay buffer from training a DQN agent (50 million transitions), it is clear there is a reliance on their specific setting for these algorithms to perform well. We do remark that our results confirm their observation that distributional reinforcement learning algorithms (QR-DQN) outperforms their standard counterpart (DQN), suggesting that learning a distribution aids in exploitation.

In comparison to the off-policy algorithms, the batch reinforcement learning algorithms perform reasonably well. BCQ, in particular, outperforms every other method in all tested games. However, these results also shown the current downsides of these methods. Although BCQ has the strongest performance, on most games it only matches the performance of the online DQN, which is the underlying noise-free behavioral policy. These results suggest BCQ achieves something closer to robust imitation, rather than true batch reinforcement learning when there is limited exploratory data. 
KL-Control often demonstrates a strong initial performance before failing. Examining Figure \ref{fig:values}, the drop in performance corresponds to a negative divergence in the value estimate. In the games where the value estimate does not diverge (Enduro and Seaquest), KL-Control performs well. It is possible that with additional hyper-parameter tuning, stability could be maintained, however, this suggests that KL-Control is not robust to hyper-parameters or varied tasks.

These results additionally confirm the experiments from \cite{fujimoto2019off}, which showed that standard off-policy deep reinforcement learning algorithms fail in the batch setting, due to high extrapolation error from selecting out-of-distribution actions during value updates. 
Furthermore, it is clear that the algorithms with the strongest performance also have stable value estimates, suggesting that the mitigation of extrapolation error is important for batch deep reinforcement learning.

\section{Conclusion}

In this paper, we perform empirical analysis on current off-policy and batch reinforcement learning algorithms in a simple single behavioral policy task on several Atari environments \citep{bellemare2013arcade}. Our experiments show that current algorithms fail to achieve a satisfactory performance in this setting, by under-performing online DQN and the behavioral policy. Our results suggest that algorithms which do not consider extrapolation error or the distribution of data will perform poorly in the batch setting with low data diversity, due to unstable value estimates. Lastly, we introduce a discrete version of Batch-Constrained deep Q-learning (BCQ) \citep{fujimoto2019off}, which outperforms all previous algorithms in this setting, while being straightforward to implement. We hope BCQ will serve as a strong baseline for future methods in this area. 
\vfill

\bibliographystyle{plainnat}
\bibliography{references}

\clearpage

\title{Supplementary Material}
\maketitle

\appendix

\section{Experimental Details}

\subsection{Atari Preprocessing.}

The Atari 2600 environment is preprocessed in the same manner as previous work \citep{DQN, machado2018revisiting, castro2018dopamine} and we use consistent preprocessing across all tasks and algorithms. 

We denote the output of the Atari environment as \textit{frames}. These frames are grayscaled and resized to $84$ by $84$ pixels. Furthermore, the agent only receives a \textit{state} and selects an action every $4$th frame. The selected action is repeated for the next $4$ frames. The state is defined by the maximum between the previous two frames. Furthermore, the input to the networks is a concatenation of the previous $4$ states. This means each network receives a tensor with dimensions $(4, 84, 84)$, which considers a history of $16$ frames ($4$ frames over $4$ states). For the first $3$ time steps, the input to the networks includes \textit{states} which are set to all $0$s. If the environment terminates before $4$ frames have passed, the state is defined by the maximum of the final two frames before termination. 

In accordance to \cite{machado2018revisiting}, sticky actions are used, such that the action $a_t$ is set to the previous action $a_{t-1}$ with probability $p=0.25$. No-operations are not applied at the beginning of episodes, and random frame skips are not used.

The reward function is defined by the in-game reward, but clipped to a range of $[-1, 1]$. The environment terminates when the game terminates (rather than on a lost life), or after $27$k time steps, corresponding to $108$k frames or $30$ minutes of real time. 

\subsection{Architecture and Hyper-parameters.}

Unless stated otherwise, all networks use the same architecture and hyper-parameters.

Image inputs are passed through a 3-layered convolutional neural network taking an input size of $(4, 84, 84)$. The first layer has a kernel depth of $32$, kernel size of $8 \times 8$ and stride $4$. The second layer has a kernel depth of $32$, kernel size of $4 \times 4$ and stride $2$. The third layer has a kernel depth of $64$, kernel size of $2 \times 2$ and stride $1$. This output is flattened into a vector of $3136$ and passed to a full-connected network with one hidden layer of $512$. The output of the Q-network is a Q-value for each action. ReLU activation functions are used after layer besides the final fully-connected layer. 

For methods which require a generative model, the convolutional neural network is shared between both the Q-network and the generative model. The generative model is a secondary fully-connected network with the same architecture. The final layer uses a softmax activation after the output of the network, to recover probabilities for each action. 

Hyper-parameters are held consistent across each algorithm and listed in Table \ref{table:hyperparameters}. Hyper-parameters were chosen to match the implementation of Rainbow \citep{hessel2017rainbow} in the Dopamine framework~\citep{castro2018dopamine}.

\begin{table}[ht]
\centering
\caption{Hyper-parameters used by each network.}
\begin{tabular}{l|l}
\hline
Hyper-parameter & Value\\ 
\hline
Network optimizer & Adam \citep{adam}\\
Learning rate & $0.0000625$\\
Adam $\e$ & $0.00015$\\
Discount $\y$ & $0.99$\\
Mini-batch size & $32$\\
Target network update frequency & $8$k training iterations\\
Huber loss $\kappa$ & $1$\\
Evaluation $\e$ & $0.001$\\
\hline
\end{tabular}
\label{table:hyperparameters}
\end{table}

Algorithm-specific hyper-parameters are listed in Table \ref{table:algohyperparameters}. Additionally, we regularize the generative model $G_\w$ used in BCQ and KL-Control by a penalty on the final pre-activation output $x$ by $0.01 x^2$. For KL-Control, dropout is applied before both of the fully-connected layers.

\begin{table}[ht]
\centering
\caption{Algorithm-specific hyper-parameters.}
\begin{tabular}{l|l|l}
\hline
Algorithm & Hyper-parameter & Value\\ 
\hline
QR-DQN & Quantiles $K$ & $50$\\
\hline
REM & Heads $K$ & $200$\\
\hline
BCQ & Threshold $\tau$ & $0.3$\\
\hline
\multirow{4}{*}{KL-Control} & Dropout masks $K$ & $5$\\
& KL weighting $c$ & $2$\\ 
& Gradient clipping & $1.0$\\
& Dropout probability & $0.2$\\
\hline
\end{tabular}
\label{table:algohyperparameters}
\end{table}

Additionally, we list the hyper-parameters of the online DQN, which served as the behavioral policy, in Table \ref{table:behavioral}. Training frequency corresponds to how often a training update was performed. Warmup time steps defines the initial period where actions are randomly selected and stored in the replay buffer, before any training occurs. We use $\e$-greedy for exploration, where $\e$ is decayed over time. 

\begin{table}[ht]
\centering
\caption{Hyper-parameters used by online DQN.}
\begin{tabular}{l|l}
\hline
Hyper-parameter & Value\\ 
\hline
Replay buffer size & $1$ million \\
Training frequency & Every $4$th time step\\
Warmup time steps & $20$k time steps\\
Initial $\e$ & $1.0$\\
Final $\e$ & $0.01$\\
$\e$ decay period & $250$k training iterations\\
\hline
\end{tabular}
\label{table:behavioral}
\end{table}




\end{document}